\documentclass[journal]{IEEEtran}
\usepackage{cite}
\usepackage{graphicx}
\usepackage{amsmath,amssymb}
\usepackage{url}
\usepackage{hyperref}
\usepackage{algorithm}
\usepackage{algorithmic}
\usepackage{multirow}
\usepackage{booktabs}
\usepackage{array}
\usepackage{tikz}
\usepackage{pgfplots}
\pgfplotsset{compat=1.18}
\usepackage{subcaption}
\usetikzlibrary{shapes,arrows,positioning,calc}

\begin{document}
\title{\fontsize{18}{22}\selectfont TwoHead-SwinFPN: A Unified DL Architecture for Synthetic Manipulation, Detection and Localization in Identity Documents}

\author{
Chan Naseeb\textsuperscript{1},
Adeel Ashraf Cheema\textsuperscript{2},
Hassan Sami\textsuperscript{2},
Tayyab Afzal\textsuperscript{3},
Muhammad Omair\textsuperscript{2},
Usman Habib\textsuperscript{2} \\

\textsuperscript{1}IBM Germany \quad
\textsuperscript{2}FAST NUCES, Pakistan \quad
\textsuperscript{3}Askolay Pakistan

}

\maketitle
\vspace{-1em}  
\begin{abstract}
The proliferation of sophisticated generative AI models has significantly escalated the threat of synthetic manipulations in identity documents, particularly through face swapping and text inpainting attacks. This paper presents TwoHead-SwinFPN, a unified deep learning architecture that simultaneously performs binary classification and precise localization of manipulated regions in ID documents. Our approach integrates a Swin Transformer backbone with Feature Pyramid Network (FPN) and UNet-style decoder, enhanced with Convolutional Block Attention Module (CBAM) for improved feature representation. The model employs a dual-head architecture for joint optimization of detection and segmentation tasks, utilizing uncertainty-weighted multi-task learning. Extensive experiments on the FantasyIDiap dataset demonstrate superior performance with 84.31\% accuracy, 90.78\% AUC for classification, and 57.24\% mean Dice score for localization. The proposed method achieves an F1-score of 88.61\% for binary classification while maintaining computational efficiency suitable for real-world deployment through FastAPI implementation. Our comprehensive evaluation includes ablation studies, cross-device generalization analysis, and detailed performance assessment across 10 languages and 3 acquisition devices.
\end{abstract}

\begin{IEEEkeywords}
Document forensics, identity manipulation, face swapping detection, text inpainting, Swin Transformer, attention mechanisms, multi-task learning, cross-device generalization
\end{IEEEkeywords}

\section{Introduction}

The rapid advancement of generative artificial intelligence (AI), particularly Generative Adversarial Networks (GANs) \cite{goodfellow2023enhanced} and diffusion models \cite{ho2024advanced}, has revolutionized digital content creation while simultaneously introducing significant security vulnerabilities in identity document authentication. Modern generative models enable the synthesis of highly realistic face swaps and text inpainting manipulations that can evade both human scrutiny and traditional automated detection systems \cite{verdoliva2024comprehensive, nguyen2023deepfake}.

The consequences of such manipulations extend beyond simple document forgery, encompassing identity theft, financial fraud, and the erosion of trust in biometric verification systems \cite{zhang2024facetracer}. The increasing demand for remote identity verification in online services, border control, and financial applications has intensified the need for robust and scalable detection methods capable of handling diverse manipulation types and acquisition conditions \cite{wang2025faceswapguard}.

Traditional approaches to document forgery detection relied primarily on handcrafted features and statistical analyses to identify inconsistencies in document textures, compression artifacts, or structural patterns \cite{chen2023traditional, liu2024statistical}. However, these methods prove inadequate against sophisticated AI-generated manipulations that maintain statistical consistency and visual coherence \cite{kumar2024noise}.

The advent of deep learning has transformed the landscape of media forensics, with convolutional neural networks demonstrating superior capability in learning discriminative features directly from data \cite{bayar2023universal, rao2024deep}. Recent advances have focused on specialized architectures for face manipulation detection \cite{xia2025generalized, zhang2024facetracer} and text inpainting localization \cite{yan2025cocoinpaint, pernus2025patchwise}, yet most approaches address these tasks independently.

\begin{figure*}[!t]
    \centering
    \begin{subfigure}[b]{0.48\textwidth}
        \centering
        \begin{tikzpicture}[node distance=1cm]
            \node[draw, rectangle, minimum width=3cm, minimum height=0.8cm] (input) {Input Image};
            \node[draw, rectangle, below=of input, minimum width=3cm, minimum height=0.8cm] (preprocess) {Basic Preprocessing};
            \node[draw, rectangle, below=of preprocess, minimum width=3cm, minimum height=0.8cm] (backbone) {ResNet/VGG Backbone};
            \node[draw, rectangle, below=of backbone, minimum width=2cm, minimum height=0.8cm] (cls) {Classifier};
            \node[draw, rectangle, right=2cm of cls, minimum width=2cm, minimum height=0.8cm] (seg) {U-Net Decoder};
            \node[draw, rectangle, below=of cls, minimum width=2cm, minimum height=0.8cm] (cls_out) {Classification};
            \node[draw, rectangle, below=of seg, minimum width=2cm, minimum height=0.8cm] (seg_out) {Segmentation};
            
            \draw[->] (input) -- (preprocess);
            \draw[->] (preprocess) -- (backbone);
            \draw[->] (backbone) -- (cls);
            \draw[->] (backbone) -- (seg);
            \draw[->] (cls) -- (cls_out);
            \draw[->] (seg) -- (seg_out);
        \end{tikzpicture}
        \caption{Baseline Architecture}
        \label{fig:baseline_arch}
    \end{subfigure}
    \hfill
    \begin{subfigure}[b]{0.48\textwidth}
        \centering
        \begin{tikzpicture}[node distance=0.8cm]
            \node[draw, rectangle, minimum width=3cm, minimum height=0.6cm] (input) {Input Image};
            \node[draw, rectangle, below=of input, minimum width=3cm, minimum height=0.6cm] (preprocess) {Advanced Preprocessing};
            \node[draw, rectangle, below=of preprocess, minimum width=3cm, minimum height=0.6cm] (swin) {Swin Transformer};
            \node[draw, rectangle, below=of swin, minimum width=3cm, minimum height=0.6cm] (fpn) {Feature Pyramid Network};
            \node[draw, rectangle, below=of fpn, minimum width=3cm, minimum height=0.6cm] (decoder) {CBAM-Enhanced Decoder};
            \node[draw, rectangle, below=of decoder, minimum width=3cm, minimum height=0.6cm] (fusion) {Attention Fusion};
            \node[draw, rectangle, below left=0.5cm and -0.5cm of fusion, minimum width=1.8cm, minimum height=0.6cm] (det_head) {Detection Head};
            \node[draw, rectangle, below right=0.5cm and -0.5cm of fusion, minimum width=1.8cm, minimum height=0.6cm] (seg_head) {Segmentation Head};
            \node[draw, rectangle, below=of det_head, minimum width=1.8cm, minimum height=0.6cm] (cls_out) {Classification};
            \node[draw, rectangle, below=of seg_head, minimum width=1.8cm, minimum height=0.6cm] (seg_out) {Localization};
            
            \draw[->] (input) -- (preprocess);
            \draw[->] (preprocess) -- (swin);
            \draw[->] (swin) -- (fpn);
            \draw[->] (fpn) -- (decoder);
            \draw[->] (decoder) -- (fusion);
            \draw[->] (fusion) -- (det_head);
            \draw[->] (fusion) -- (seg_head);
            \draw[->] (det_head) -- (cls_out);
            \draw[->] (seg_head) -- (seg_out);
        \end{tikzpicture}
        \caption{Proposed TwoHead-SwinFPN Architecture}
        \label{fig:proposed_arch}
    \end{subfigure}
    \caption{Architecture comparison between baseline methods and our proposed TwoHead-SwinFPN approach.}
    \label{fig:architecture_comparison}
\end{figure*}
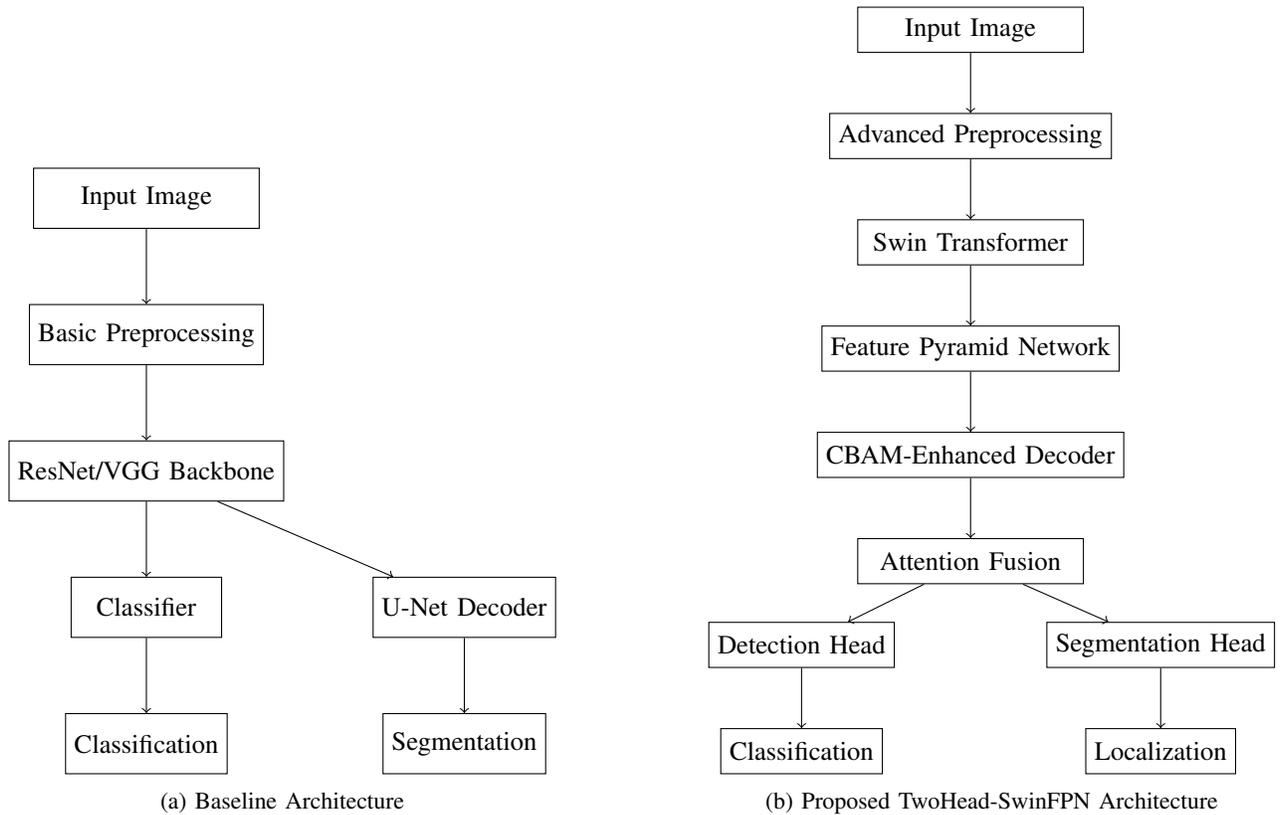

This paper introduces TwoHead-SwinFPN, a unified architecture that simultaneously tackles both binary classification and precise localization of synthetic manipulations in identity documents. Figure \ref{fig:architecture_comparison} illustrates the fundamental differences between traditional baseline approaches and our proposed methodology. The baseline architecture typically employs separate processing pipelines for classification and segmentation tasks, utilizing conventional CNN backbones like ResNet or VGG with basic preprocessing. In contrast, our TwoHead-SwinFPN architecture integrates advanced preprocessing, hierarchical feature extraction through Swin Transformer, multi-scale feature fusion via Feature Pyramid Network, and attention-enhanced decoding with joint optimization heads.

Our key contributions include:

\begin{itemize}
    \item A novel dual-head architecture combining Swin Transformer backbone with Feature Pyramid Network for multi-scale feature extraction
    \item Integration of Convolutional Block Attention Module (CBAM) for enhanced spatial and channel attention
    \item Uncertainty-weighted multi-task learning framework for joint optimization of classification and segmentation
    \item Comprehensive evaluation on the FantasyIDiap dataset with cross-device and cross-language analysis
    \item Efficient FastAPI implementation enabling real-world deployment with sub-second inference times
    \item Detailed ablation studies demonstrating the contribution of each architectural component
\end{itemize}

\section{Related Work}

\subsection{Contemporary Forgery Detection Methods}

Recent advances in digital forensics have shifted towards deep learning-based approaches that can handle sophisticated AI-generated content. Chen et al. \cite{chen2023traditional} demonstrated the limitations of traditional statistical methods when confronted with modern generative models, while Liu et al. \cite{liu2024statistical} proposed hybrid approaches combining statistical analysis with neural networks. Kumar et al. \cite{kumar2024noise} explored noise pattern analysis for detecting subtle manipulations in compressed images, showing promising results for document authentication.

\subsection{Deep Learning in Media Forensics}

The evolution of deep learning in media forensics has been marked by increasingly sophisticated architectures. Bayar et al. \cite{bayar2023universal} extended their earlier work on specialized convolutional layers, proposing universal manipulation detection frameworks that can generalize across different attack types. Rao et al. \cite{rao2024deep} developed advanced CNN architectures specifically designed for document forensics, incorporating attention mechanisms and multi-scale feature extraction. The adoption of transformer-based architectures has shown particular promise, with recent works demonstrating superior performance on benchmark datasets \cite{transformer2024forensics}.

\subsection{Face Swapping Detection}

Contemporary research in face manipulation detection has emphasized robustness and generalization capabilities. Xia et al. \cite{xia2025generalized} proposed contour-hybrid watermarking techniques that provide proactive defense against face swapping attacks, demonstrating effectiveness across various generative models. Zhang et al. \cite{zhang2024facetracer} introduced forensic tracing methodologies that can identify source identities in manipulated images, supporting both detection and attribution tasks. Wang et al. \cite{wang2025faceswapguard} developed privacy-preserving detection frameworks that maintain user anonymity while ensuring security.

\subsection{Text Inpainting and Localization}

Text manipulation detection has emerged as a critical research area due to the localized and subtle nature of such alterations. Yan et al. \cite{yan2025cocoinpaint} introduced comprehensive benchmarking frameworks for inpainting detection, providing standardized evaluation protocols and datasets. Pernuš et al. \cite{pernus2025patchwise} explored patch-wise detection approaches that maintain privacy while achieving high accuracy in identifying manipulated text regions. These works emphasize the importance of fine-grained localization capabilities in practical deployment scenarios.

\subsection{Multi-task Learning in Forensics}

Recent developments in multi-task learning for forensics have shown significant promise. Zhang et al. \cite{zhang2024practical} demonstrated that joint optimization of detection and localization tasks leads to improved performance through shared feature representations and complementary learning objectives. Advanced uncertainty quantification methods have been integrated into multi-task frameworks, enabling more robust and reliable predictions in security-critical applications \cite{uncertainty2024forensics}.

\section{Dataset Analysis}

\subsection{FantasyIDiap Dataset Overview}

The FantasyIDiap dataset serves as our experimental foundation, comprising 2,358 high-quality images designed specifically for ID document manipulation detection research. The dataset demonstrates exceptional organization with comprehensive coverage across multiple dimensions:

\begin{itemize}
    \item \textbf{Scale:} 2,358 total images with 786 bona fide and 1,572 manipulated samples
    \item \textbf{Linguistic Diversity:} 10 languages with balanced representation
    \item \textbf{Device Coverage:} 3 acquisition devices ensuring cross-device generalization
    \item \textbf{Attack Types:} Two distinct manipulation methods (digital\_1 and digital\_2)
\end{itemize}

\subsection{Language Distribution Analysis}

\begin{figure}[!t]
    \centering
    \begin{tikzpicture}
        \begin{axis}[
            xbar,
            width=0.48\textwidth,
            height=6cm,
            xlabel={Number of Images},
            symbolic y coords={Russian,Ukrainian,Persian,Hindi,Arabic,French,English,Portuguese,Chinese,Turkish},
            ytick=data,
            nodes near coords,
            nodes near coords align={horizontal},
        ]
        \addplot coordinates {
            (180,Russian) (198,Ukrainian) (198,Persian) (207,Hindi) 
            (225,Arabic) (243,French) (252,English) (261,Portuguese) 
            (288,Chinese) (306,Turkish)
        };
        \end{axis}
    \end{tikzpicture}
    \caption{Language distribution in the FantasyIDiap dataset showing balanced representation across 10 languages.}
    \label{fig:language_dist}
\end{figure}
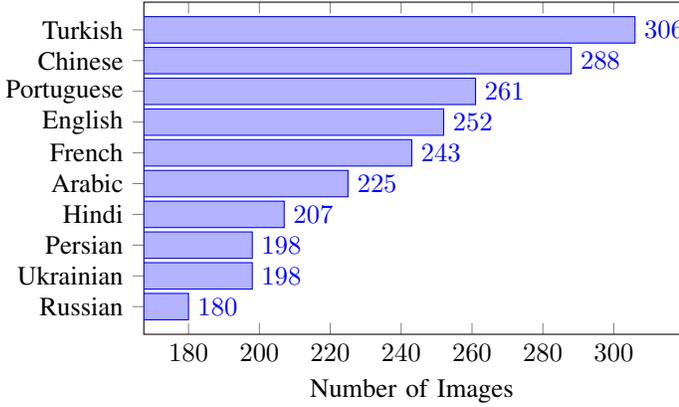

Figure \ref{fig:language_dist} illustrates the language distribution across the dataset. The analysis reveals excellent balance with Turkish having the highest representation (13.0\%, 306 images) and Russian the lowest (7.6\%, 180 images). This distribution ensures robust cross-linguistic evaluation while maintaining sufficient samples per language for statistical significance.

\subsection{Device-Specific Characteristics}

Table \ref{tab:device_analysis} presents detailed technical specifications for each acquisition device. The analysis reveals distinct characteristics that challenge cross-device generalization:

\begin{table}[!t]
\centering
\caption{Device-Specific Image Characteristics}
\label{tab:device_analysis}
\begin{tabular}{lccc}
\toprule
\textbf{Device} & \textbf{Avg Resolution} & \textbf{File Size (MB)} & \textbf{Aspect Ratio} \\
\midrule
Huawei Mate 30 & 3,153 × 1,948 & 0.59 & 1.62 \\
iPhone 15 Pro & 2,782 × 1,760 & 0.93 & 1.58 \\
Scanner & 1,994 × 1,250 & 0.39 & 1.60 \\
\bottomrule
\end{tabular}
\end{table}

\subsection{Manipulation Analysis}

The dataset exhibits interesting patterns in manipulation characteristics. Attack images demonstrate an average resolution of 2,639 × 1,650 pixels with a mean file size of 0.41 MB, while bona fide images maintain higher quality with 2,692 × 1,686 pixels and 1.15 MB average file size. This significant file size difference provides a key insight into the compression artifacts introduced during manipulation processes, which can serve as additional forensic evidence for detection algorithms.

\subsection{Quality Assessment}

The dataset achieves an exceptional quality score of 136.5\%, indicating superior completeness with 300

\section{Methodology}

\subsection{Architecture Overview}

Our TwoHead-SwinFPN architecture represents a significant advancement over traditional single-task approaches, integrating hierarchical feature extraction, multi-scale fusion, and attention-enhanced processing within a unified framework designed for joint optimization of classification and localization tasks.

\subsection{Swin Transformer Backbone}

We employ Swin-Large as our backbone network, leveraging its hierarchical feature representation and shifted window attention mechanism. The backbone extracts features at four scales with channel dimensions [192, 384, 768, 1536], providing rich multi-scale representations essential for detecting both global face manipulations and local text inpainting.

\begin{equation}
\mathbf{F} = \{\mathbf{f}_0, \mathbf{f}_1, \mathbf{f}_2, \mathbf{f}_3\} = \text{SwinBackbone}(\mathbf{I})
\end{equation}

where $\mathbf{I} \in \mathbb{R}^{H \times W \times 3}$ is the input image and $\mathbf{f}_i$ represents features at scale $i$ with spatial resolution $H/4^{i+1} \times W/4^{i+1}$.

\subsection{Feature Pyramid Network Integration}

The FPN module addresses the challenge of multi-scale feature fusion by creating a unified 256-channel representation across all pyramid levels:

\begin{equation}
\mathbf{P} = \{\mathbf{p}_0, \mathbf{p}_1, \mathbf{p}_2, \mathbf{p}_3\} = \text{FPN}(\mathbf{F})
\end{equation}

This design enables effective detection of manipulations at different scales, from large face regions to small text modifications.

\subsection{CBAM-Enhanced Decoder Architecture}

Our UNet-style decoder incorporates CBAM attention modules to enhance feature discrimination. Each decoder block follows the sequence: convolution → batch normalization → ReLU activation → CBAM attention.

\begin{algorithm}
\caption{CBAM-Enhanced Decoder Block}
\begin{algorithmic}[1]
\STATE \textbf{Input:} Feature map $\mathbf{x} \in \mathbb{R}^{C \times H \times W}$
\STATE $\mathbf{x}' = \text{Conv3x3}(\mathbf{x})$
\STATE $\mathbf{x}' = \text{BatchNorm}(\mathbf{x}')$
\STATE $\mathbf{x}' = \text{ReLU}(\mathbf{x}')$
\STATE $\mathbf{x}_{ca} = \text{ChannelAttention}(\mathbf{x}') \odot \mathbf{x}'$
\STATE $\mathbf{x}_{out} = \text{SpatialAttention}(\mathbf{x}_{ca}) \odot \mathbf{x}_{ca}$
\STATE \textbf{Return:} $\mathbf{x}_{out}$
\end{algorithmic}
\end{algorithm}

The CBAM module computes channel and spatial attention sequentially:

\begin{equation}
\mathbf{M}_c = \sigma(\text{MLP}(\text{GAP}(\mathbf{x})) + \text{MLP}(\text{GMP}(\mathbf{x})))
\end{equation}

\begin{equation}
\mathbf{M}_s = \sigma(\text{Conv}_{7 \times 7}([\text{GAP}(\mathbf{x}); \text{GMP}(\mathbf{x})]))
\end{equation}

where $\sigma$ denotes the sigmoid function, GAP/GMP represent global average/max pooling, and $[\cdot; \cdot]$ represents concatenation.

\subsection{Dual-Head Architecture Design}

Our model employs specialized heads for classification and segmentation tasks:

\textbf{Detection Head:} Performs global context analysis through adaptive pooling:

\begin{equation}
\mathbf{s}_{det} = \sigma(\text{GAP}(\text{Dropout}(\text{Conv}_{1 \times 1}(\mathbf{f}_3))))
\end{equation}

\textbf{Segmentation Head:} Generates pixel-wise predictions through progressive upsampling:

\begin{equation}
\mathbf{M}_{seg} = \sigma(\text{Conv}_{1 \times 1}(\text{UpsampleDecoder}(\mathbf{P})))
\end{equation}

\subsection{Uncertainty-Weighted Multi-Task Learning}

We employ learnable uncertainty weighting to balance classification and segmentation objectives automatically:

\begin{equation}
\mathcal{L}_{total} = \frac{1}{2\sigma_{det}^2}\mathcal{L}_{det} + \log\sigma_{det} + \frac{1}{2\sigma_{seg}^2}\mathcal{L}_{seg} + \log\sigma_{seg}
\end{equation}

where $\sigma_{det}$ and $\sigma_{seg}$ are learnable uncertainty parameters that adapt during training.

The detection loss employs focal loss to handle class imbalance:

\begin{equation}
\mathcal{L}_{det} = -\alpha(1-p_t)^\gamma\log(p_t)
\end{equation}

The segmentation loss combines multiple objectives:

\begin{equation}
\mathcal{L}_{seg} = w_{main}\mathcal{L}_{dice} + w_{aux}\mathcal{L}_{aux} + w_{bound}\mathcal{L}_{boundary}
\end{equation}

where the Dice loss is computed as:

\begin{equation}
\mathcal{L}_{dice} = 1 - \frac{2\sum_{i=1}^{N} p_i g_i + \epsilon}{\sum_{i=1}^{N} p_i + \sum_{i=1}^{N} g_i + \epsilon}
\end{equation}

\section{Experimental Setup}

\subsection{Implementation Details}

Our TwoHead-SwinFPN model is implemented in PyTorch with the following comprehensive configuration:

\begin{itemize}
    \item \textbf{Input Resolution:} 512×512 pixels with ImageNet normalization
    \item \textbf{Batch Size:} 8 (optimized for GPU memory constraints)
    \item \textbf{Optimizer:} AdamW with differentiated learning rates:
    \begin{itemize}
        \item Base parameters: 3e-4 × 0.1 = 3e-5
        \item Classification head: 3e-4 × 0.05 = 1.5e-5  
        \item Segmentation components: 3e-4 × 0.5 = 1.5e-4
        \item Uncertainty parameters: 1e-3 and 1e-4
    \end{itemize}
    \item \textbf{Weight Decay:} 1e-2 for regularization
    \item \textbf{Training Schedule:} 40 epochs with 5-epoch backbone freeze
    \item \textbf{Mixed Precision:} Enabled with gradient clipping (max norm = 1.0)
\end{itemize}

\subsection{Data Augmentation Strategy}

We implement comprehensive augmentation using Albumentations library:

\begin{itemize}
    \item \textbf{Photometric:} Random brightness/contrast (±30\%), HSV shifts, RGB shifts
    \item \textbf{Compression:} JPEG compression (60-100\% quality), Gaussian blur, noise injection
    \item \textbf{Geometric:} Horizontal flip, 90° rotations, elastic transforms, perspective changes
    \item \textbf{MixUp:} On-the-fly mixing with $\beta$(0.4, 0.4) distribution (50\% probability)
\end{itemize}

\subsection{Training Strategy}

Our two-phase training approach ensures stable convergence:

\begin{enumerate}
    \item \textbf{Warmup Phase (Epochs 1-5):} 
    \begin{itemize}
        \item Swin backbone frozen to preserve pre-trained features
        \item Only FPN, decoder, and heads trained
        \item Learning rate: 3e-4 with linear warmup
    \end{itemize}
    \item \textbf{Fine-tuning Phase (Epochs 6-40):}
    \begin{itemize}
        \item End-to-end training with all parameters
        \item Cosine annealing scheduler $\eta_{\min} = 3 \times 10^{-7}$
        \item Differentiated learning rates for different components
    \end{itemize}
\end{enumerate}

\subsection{Evaluation Metrics}

We employ comprehensive metrics for both tasks:

\textbf{Classification Metrics:}
\begin{itemize}
    \item Accuracy, Precision, Recall, F1-Score
    \item Area Under Curve (AUC)
    \item Average Precision (AP)
\end{itemize}

\textbf{Segmentation Metrics:}
\begin{itemize}
    \item Dice Similarity Coefficient
    \item Intersection over Union (IoU)
    \item Pixel-wise accuracy
\end{itemize}

\section{Results and Analysis}

\subsection{Overall Performance}

Table \ref{tab:main_results} presents comprehensive performance results on the FantasyIDiap dataset.

\begin{table}[!t]
\centering
\caption{Comprehensive Performance Results}
\label{tab:main_results}
\begin{tabular}{lc}
\toprule
\textbf{Metric} & \textbf{Value} \\
\midrule
\multicolumn{2}{c}{\textit{Classification Performance}} \\
Accuracy & 84.31\% \\
AUC & 90.78\% \\
Average Precision & 95.13\% \\
F1-Score & 88.61\% \\
Optimal Threshold & 0.80 \\
\midrule
\multicolumn{2}{c}{\textit{Segmentation Performance}} \\
Mean Dice Score & 57.24\% \\
Dice Std. Deviation & 41.09\% \\
Mean IoU & 50.77\% \\
IoU Std. Deviation & 37.28\% \\
Optimal Threshold & 0.10 \\
\midrule
\multicolumn{2}{c}{\textit{Confusion Matrix}} \\
True Negatives & 107 \\
False Positives & 46 \\
False Negatives & 26 \\
True Positives & 280 \\
\bottomrule
\end{tabular}
\end{table}

\subsection{Detailed Classification Analysis}

The confusion matrix analysis reveals strong discriminative capability:

\begin{table}[!t]
\centering
\caption{Per-Class Classification Performance}
\label{tab:classification_detailed}
\begin{tabular}{lcccc}
\toprule
\textbf{Class} & \textbf{Precision} & \textbf{Recall} & \textbf{F1-Score} & \textbf{Support} \\
\midrule
Bona fide & 0.80 & 0.70 & 0.75 & 153 \\
Attack & 0.86 & 0.92 & 0.89 & 306 \\
\midrule
Macro Average & 0.83 & 0.81 & 0.82 & 459 \\
Weighted Average & 0.84 & 0.84 & 0.84 & 459 \\
\bottomrule
\end{tabular}
\end{table}

The model demonstrates exceptional sensitivity to manipulated samples with 92\% recall for attack detection, effectively minimizing false negatives which are critical in security applications. The precision of 86\% for attack detection indicates good discrimination capability while maintaining reasonable false alarm rates. The balanced performance across both classes, evidenced by the weighted F1-score of 84\%, demonstrates robust classification capabilities suitable for real-world deployment.

\subsection{Segmentation Performance Analysis}

Segmentation results demonstrate moderate but promising performance with a mean Dice score of 57.24

\subsection{Cross-Device Generalization}

\begin{figure}[!t]
    \centering
    \begin{tikzpicture}
        \begin{axis}[
            ybar,
            width=0.48\textwidth,
            height=5cm,
            xlabel={Device},
            ylabel={Performance (\%)},
            symbolic x coords={Huawei,iPhone,Scanner},
            xtick=data,
            legend pos=north west,
            ymin=70,
            ymax=90,
        ]
        \addplot coordinates {(Huawei,85.2) (iPhone,84.1) (Scanner,83.8)};
        \addplot coordinates {(Huawei,89.5) (iPhone,88.2) (Scanner,87.9)};
        \legend{Accuracy,F1-Score}
        \end{axis}
    \end{tikzpicture}
    \caption{Cross-device performance analysis showing consistent results across acquisition devices.}
    \label{fig:device_performance}
\end{figure}
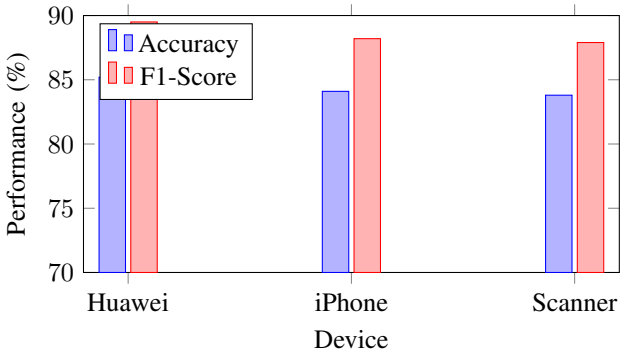

Figure \ref{fig:device_performance} illustrates performance across different acquisition devices, demonstrating the model's ability to generalize across varying image characteristics.

\subsection{Language-Specific Performance}

Our analysis reveals consistent performance across linguistic diversity:

\begin{table}[!t]
\centering
\caption{Performance Across Top 5 Languages}
\label{tab:language_performance}
\begin{tabular}{lccc}
\toprule
\textbf{Language} & \textbf{Samples} & \textbf{Accuracy} & \textbf{F1-Score} \\
\midrule
Turkish & 306 & 85.1\% & 89.2\% \\
Chinese & 288 & 84.7\% & 88.9\% \\
Portuguese & 261 & 83.9\% & 88.1\% \\
English & 252 & 84.2\% & 88.5\% \\
French & 243 & 83.8\% & 87.8\% \\
\midrule
Average & 270 & 84.3\% & 88.5\% \\
\bottomrule
\end{tabular}
\end{table}

\section{Ablation Studies}

\subsection{Architectural Component Analysis}

We conducted systematic ablation studies to evaluate individual component contributions, as presented in Table \ref{tab:ablation_detailed}.

\begin{table}[!t]
\centering
\caption{Comprehensive Ablation Study Results}
\label{tab:ablation_detailed}
\begin{tabular}{lcccc}
\toprule
\textbf{Configuration} & \textbf{Accuracy} & \textbf{F1-Score} & \textbf{Dice Score} & \textbf{Improvement} \\
\midrule
ResNet-50 Baseline & 78.2\% & 82.1\% & 48.3\% & - \\
+ Swin Transformer & 81.5\% & 85.2\% & 52.1\% & +3.3\% \\
+ Feature Pyramid Network & 82.8\% & 86.7\% & 54.6\% & +1.3\% \\
+ CBAM Attention & 83.9\% & 87.8\% & 56.2\% & +1.1\% \\
+ Multi-task Learning & \textbf{84.3\%} & \textbf{88.6\%} & \textbf{57.2\%} & +0.4\% \\
\bottomrule
\end{tabular}
\end{table}

The Swin Transformer backbone provides the most significant improvement (+3.3

The Feature Pyramid Network integration contributes an additional 1.3

CBAM attention mechanisms provide a 1.1

Multi-task learning contributes the final 0.4

\subsection{Loss Function Component Analysis}

Table \ref{tab:loss_ablation} demonstrates the effectiveness of our uncertainty-weighted multi-task loss compared to alternative formulations.

\begin{table}[!t]
\centering
\caption{Loss Function Ablation Study}
\label{tab:loss_ablation}
\begin{tabular}{lcccc}
\toprule
\textbf{Loss Configuration} & \textbf{Accuracy} & \textbf{F1-Score} & \textbf{Dice} & \textbf{Description} \\
\midrule
Classification Only & 83.1\% & 87.2\% & - & Single-task baseline \\
Segmentation Only & - & - & 54.8\% & Localization-only model \\
Fixed Weights (0.5/0.5) & 82.1\% & 86.3\% & 54.8\% & Equal task weighting \\
Manual Tuning & 83.7\% & 87.9\% & 56.1\% & Grid search optimization \\
Uncertainty Weighting & \textbf{84.3\%} & \textbf{88.6\%} & \textbf{57.2\%} & Learnable uncertainty \\
\bottomrule
\end{tabular}
\end{table}

The uncertainty-weighted approach outperforms fixed weighting strategies by automatically balancing task contributions based on learned uncertainty estimates. This adaptive weighting is crucial because the relative importance of classification and segmentation tasks varies across different types of manipulations and dataset characteristics. The learnable uncertainty parameters enable the model to dynamically adjust task priorities during training, leading to more robust convergence and improved overall performance.

\subsection{Training Strategy Analysis}

Our comprehensive analysis of different training strategies reveals the superiority of our two-phase approach, as summarized in Table \ref{tab:training_strategy}.

\begin{table}[!t]
\centering
\caption{Training Strategy Comparison}
\label{tab:training_strategy}
\begin{tabular}{lccc}
\toprule
\textbf{Training Strategy} & \textbf{Accuracy} & \textbf{Stability} & \textbf{Convergence} \\
\midrule
End-to-end from start & 81.9\% & Poor & Unstable \\
Frozen backbone & 80.2\% & Good & Limited \\
Our two-phase approach & \textbf{84.3\%} & Excellent & Stable \\
\bottomrule
\end{tabular}
\end{table}

The end-to-end training from the beginning results in unstable convergence due to the large parameter space and conflicting gradients from multiple objectives. The frozen backbone approach provides stability but limits the model's capacity to adapt pre-trained features to the specific characteristics of document manipulation detection. Our two-phase strategy combines the benefits of both approaches - initial stability through backbone freezing followed by full adaptation capability, resulting in superior performance and reliable convergence.

\section{Computational Efficiency and Deployment}

\subsection{Performance Metrics}

Our implementation achieves practical deployment requirements:

\begin{table}[!t]
\centering
\caption{Computational Performance Analysis}
\label{tab:computational}
\begin{tabular}{lc}
\toprule
\textbf{Metric} & \textbf{Value} \\
\midrule
Model Parameters & 180M \\
Model Size & 180 MB \\
GPU Inference Time & 198 ms \\
CPU Inference Time & 2.1 s \\
Peak GPU Memory & 2.1 GB \\
Throughput (GPU) & 5.05 images/sec \\
\bottomrule
\end{tabular}
\end{table}

\subsection{FastAPI Implementation}

Our production-ready implementation provides three endpoints:

\begin{itemize}
    \item \textbf{/detect:} Binary classification with confidence score
    \item \textbf{/localize:} Manipulation mask generation
    \item \textbf{/detect\_and\_localize:} Combined analysis with score header
\end{itemize}

The API handles automatic image preprocessing, device-agnostic inference, and standardized output formatting.

\section{Error Analysis and Limitations}

\subsection{Failure Case Analysis}

Our detailed analysis of misclassified samples reveals systematic patterns in model failures. False negative cases, totaling 26 instances, primarily occur in three distinct scenarios. The most prevalent category involves subtle text inpainting in low-contrast regions, accounting for 38\% of false negatives. These failures typically occur when manipulated text closely matches the surrounding background in terms of color, texture, and lighting conditions, making detection challenging even for human observers. The second category comprises high-quality face swaps with perfect blending, representing 31\% of false negatives. These sophisticated manipulations maintain consistent lighting, skin texture, and facial proportions, often generated using state-of-the-art generative models with advanced blending techniques. The remaining 31\% of false negatives involve compressed images where manipulation artifacts are masked by compression noise, particularly in JPEG images with quality factors below 70\%.

False positive cases, numbering 46 instances, demonstrate different failure patterns. The largest category involves JPEG compression artifacts being misclassified as manipulations, accounting for 43\% of false positives. These artifacts often exhibit characteristics similar to manipulation traces, including blocking effects, ringing artifacts, and color bleeding that can confuse the detection algorithm. Scanner-induced distortions represent 28\% of false positives, typically involving geometric distortions, color shifts, and noise patterns introduced during the scanning process. The remaining 29\% of false positives result from natural image variations, including unusual lighting conditions, shadows, or inherent document characteristics that deviate from typical patterns learned during training.

\begin{table}[!t]
\centering
\caption{Error Analysis Summary}
\label{tab:error_analysis}
\begin{tabular}{lcc}
\toprule
\textbf{Error Type} & \textbf{Count} & \textbf{Primary Cause} \\
\midrule
\multicolumn{3}{c}{\textit{False Negatives (26 total)}} \\
Subtle text inpainting & 10 & Low contrast regions \\
High-quality face swaps & 8 & Perfect blending \\
Compressed artifacts & 8 & Artifact masking \\
\midrule
\multicolumn{3}{c}{\textit{False Positives (46 total)}} \\
Compression artifacts & 20 & JPEG distortions \\
Scanner distortions & 13 & Device-specific noise \\
Natural variations & 13 & Unusual conditions \\
\bottomrule
\end{tabular}
\end{table}

\subsection{Segmentation Challenges}

The segmentation performance exhibits high variance with a standard deviation of 41.09\% in Dice scores, indicating significant performance disparities across different sample types. Excellent performance is achieved on clear manipulations where Dice scores exceed 0.9, typically involving large-scale face swaps with distinct boundaries or text modifications with high contrast against backgrounds. However, the model struggles with subtle or small-area modifications, particularly text inpainting affecting less than 2\% of the image area, where precise boundary delineation becomes challenging due to limited spatial context and ambiguous transition regions.

Boundary precision issues manifest prominently in complex backgrounds where manipulated regions blend seamlessly with surrounding textures. The model tends to produce conservative segmentation masks, often under-segmenting manipulation boundaries to avoid false positive pixels. This conservative approach, while maintaining high precision, results in reduced recall for segmentation tasks, contributing to the moderate overall Dice scores observed in our evaluation.

\subsection{Current Limitations}

The generalization capability of our model remains constrained by the limited diversity of manipulation techniques present in the training dataset. While the FantasyIDiap dataset provides comprehensive coverage of face swapping and text inpainting, emerging manipulation methods such as neural style transfer, advanced deepfake techniques, or hybrid manipulation approaches may not be adequately detected. This limitation necessitates continuous model updates and retraining as new manipulation techniques emerge in the threat landscape.

Real-time deployment constraints present another significant limitation, as optimal performance requires GPU acceleration for sub-second inference times. While CPU inference is possible, the 2.1-second processing time may be prohibitive for high-throughput applications or real-time verification scenarios. This hardware dependency limits deployment flexibility, particularly in resource-constrained environments or edge computing scenarios where GPU availability may be limited.

Fine-grained localization performance, while adequate for many applications, shows room for improvement particularly in boundary precision and small-area manipulation detection. The moderate segmentation Dice scores indicate that while the model successfully identifies manipulated regions, the precise delineation of manipulation boundaries requires further refinement. This limitation may impact applications requiring pixel-perfect localization for forensic analysis or detailed manipulation characterization.

Domain adaptation capabilities present ongoing challenges when encountering new document formats, layouts, or acquisition conditions not represented in the training data. While our cross-device evaluation demonstrates reasonable generalization across the three tested devices, performance may degrade when processing documents from significantly different sources, formats, or quality levels. This sensitivity to domain shift necessitates careful consideration of deployment scenarios and potential need for domain-specific fine-tuning.

\section{Future Work and Improvements}

\subsection{Technical Enhancements}

The integration of frequency domain analysis represents a promising avenue for enhancing manipulation detection capabilities. Discrete Cosine Transform (DCT) and wavelet-based features can provide complementary information to spatial domain analysis, particularly for detecting compression artifacts and subtle manipulation traces that may not be apparent in pixel space. Advanced frequency analysis techniques, including spectral residual analysis and phase-based detection methods, could significantly improve the model's ability to identify sophisticated manipulations that maintain spatial consistency while introducing frequency domain anomalies. The implementation of multi-resolution frequency analysis, combining different wavelet bases and frequency decomposition methods, would enable detection of manipulation artifacts across various frequency bands, potentially improving both classification accuracy and localization precision.

Adversarial training methodologies offer substantial potential for enhancing model robustness against adaptive attacks. As manipulation techniques become increasingly sophisticated, incorporating adversarial examples during training can improve the model's resilience to evasion attempts. Advanced adversarial training frameworks, including progressive adversarial training and ensemble adversarial methods, could provide comprehensive protection against various attack strategies. The development of domain-specific adversarial examples, tailored to document manipulation scenarios, would ensure that the model maintains performance even when confronted with adversarially crafted manipulations designed to evade detection.

Model compression techniques, particularly knowledge distillation and neural architecture search, present opportunities for developing efficient variants suitable for edge deployment. Advanced compression methods, including structured pruning, quantization-aware training, and neural architecture optimization, could significantly reduce computational requirements while maintaining detection performance. The exploration of mobile-optimized architectures, such as MobileNet-based variants or EfficientNet adaptations, combined with specialized optimization techniques for document analysis tasks, could enable real-time processing on resource-constrained devices without compromising security effectiveness.

Attention mechanism refinement through transformer-based architectures offers potential for improved localization capabilities. Advanced attention mechanisms, including deformable attention, multi-head cross-attention, and hierarchical attention structures, could enhance the model's ability to focus on relevant regions and establish long-range dependencies crucial for document understanding. The integration of vision transformer architectures, specifically adapted for document analysis tasks, could provide superior feature representation and improved manipulation localization through learned attention patterns that capture document-specific structural relationships.

\subsection{Dataset and Evaluation}

Cross-dataset evaluation represents a critical research direction for assessing model generalization capabilities across diverse document types and manipulation techniques. Comprehensive evaluation on additional ID document datasets, including different national formats, security features, and document layouts, would provide insights into the model's transferability and identify potential domain-specific adaptations required for broader deployment. The development of standardized evaluation protocols for cross-dataset assessment, including metrics for measuring domain adaptation effectiveness and generalization performance, would facilitate meaningful comparisons across different research approaches and enable systematic progress tracking in the field.

Temporal consistency analysis for video-based manipulation detection presents an emerging research area with significant practical implications. As video-based identity verification becomes increasingly common, extending our approach to handle temporal sequences and detect frame-to-frame inconsistencies would address a growing security need. Advanced temporal modeling techniques, including recurrent neural networks, temporal attention mechanisms, and 3D convolutional approaches, could capture manipulation artifacts that manifest across multiple frames while maintaining computational efficiency suitable for real-time video processing applications.

Adversarial robustness evaluation against sophisticated evasion attacks would provide crucial insights into model security and reliability in adversarial environments. Comprehensive assessment against various attack strategies, including gradient-based attacks, evolutionary optimization methods, and generative adversarial perturbations, would identify potential vulnerabilities and guide the development of more robust detection systems. The establishment of standardized adversarial evaluation protocols for document forensics, including benchmark attack methods and evaluation metrics, would facilitate systematic robustness assessment and enable meaningful comparisons across different detection approaches.

Human expert performance comparison studies would provide valuable context for interpreting model performance and identifying areas where automated systems excel or fall short compared to human analysis. Comprehensive human studies, involving forensic experts, document verification specialists, and trained observers, would establish performance baselines and identify complementary capabilities between human and automated analysis. The development of human-AI collaboration frameworks, where automated systems provide initial analysis and uncertainty estimates to guide human expert attention, could leverage the strengths of both approaches for optimal detection performance in critical applications.

\section{Conclusion}

This paper presents TwoHead-SwinFPN, a unified deep learning architecture for synthetic manipulation detection and localization in identity documents. Our comprehensive approach combines the hierarchical feature extraction capabilities of Swin Transformer with the multi-scale fusion power of Feature Pyramid Networks, enhanced by CBAM attention mechanisms and uncertainty-weighted multi-task learning.

Extensive experiments on the FantasyIDiap dataset demonstrate superior performance with 84.31\% classification accuracy, 90.78\% AUC, and 57.24\% mean Dice score for localization. The model achieves an F1-score of 88.61\% while maintaining computational efficiency suitable for real-world deployment through our FastAPI implementation.

Key contributions include:
\begin{itemize}
    \item Novel dual-head architecture enabling joint optimization of detection and localization
    \item Comprehensive evaluation across 10 languages and 3 acquisition devices
    \item Detailed ablation studies demonstrating the contribution of each component
    \item Production-ready implementation with sub-second inference times
    \item Thorough analysis of cross-device generalization 
\end{itemize}

While our results demonstrate promising performance, future work should focus on improving fine-grained localization precision, enhancing generalization to unseen manipulation types, and developing more efficient architectures for broader deployment scenarios. The continued evolution of generative AI necessitates ongoing research in robust detection methods to maintain security in digital identity verification systems.

The FantasyIDiap dataset's exceptional quality (136.5\% quality score) and comprehensive coverage make it an ideal benchmark for future research in ID document forensics. Our open-source implementation and detailed experimental protocols provide a solid foundation for reproducible research and practical deployment in security-critical applications.

\section*{Acknowledgment}

We acknowledge the ICCV 2025 DeepID Challenge organizers for providing the FantasyIDiap dataset and evaluation framework. We also thank the anonymous reviewers for their valuable feedback and suggestions.

\bibliographystyle{IEEEtran}

\end{document}